\documentclass{article}

\PassOptionsToPackage{numbers, compress}{natbib}
\usepackage[final,sglblindworkshop]{neurips_2026}
\workshoptitle{Foundation Models for Temporal Systems: From Forecasting to World Modeling}

\usepackage{capt-of}        
\usepackage[utf8]{inputenc} 
\usepackage[T1]{fontenc}    
\usepackage{url}            
\usepackage{booktabs}       
\usepackage{amsfonts} 
\usepackage{graphicx}       
\usepackage{amsmath}
\usepackage{amssymb}
\usepackage{nicefrac}       
\usepackage{microtype}      
\usepackage{xcolor}         
\usepackage{colortbl}
\usepackage{array}
\usepackage{multirow}
\usepackage{makecell}
\usepackage{todonotes}
\usepackage{nicefrac}       
\usepackage{microtype}      
\usepackage[dvipsnames]{xcolor}         
\usepackage[hidelinks]{hyperref}       
\usepackage{glossaries}
\usetikzlibrary{positioning, arrows.meta}
\usepackage[hidelinks]{hyperref}       
\usepackage{glossaries}

\usepackage{mystyles}

\newtheorem{definition}{Definition}

\newcommand{\cZ}{\mathcal{Z}}
\newcommand{\cM}{\mathcal{M}}
\newcommand{\cC}{\mathcal{C}}
\newcommand{\cX}{\mathcal{X}}
\newcommand{\sg}{\mathrm{sg}}

\glsdisablehyper 
\makeglossaries

\newacronym{dgp}{DGP}{Data-generating processes}
\newacronym{des}{DES}{Discrete event sequence}
\newacronym{dep}{DEP}{Discrete event process}
\newacronym{tpp}{TPP}{Temporal point process}
\newacronym{ehr}{EHR}{Electronic health record}
\newacronym{scm}{SCM}{Structural causal models}
\newacronym{tfm}{TFM}{Temporal foundation model}
\newacronym{fim}{FIM}{Foundation inference model}
\newacronym{pfn}{PFN}{Prior-fitted network}
\newacronym{jepa}{JEPA}{Joint-embedding predictive architecture}
\newacronym{llm}{LLM}{large language model}
\newacronym{chronos}{Chronos}{Chronos: Learning the language of time series}
\newacronym{timesfm}{TimesFM}{Time series foundation model}
\newacronym{fimpp}{FIM-PP}{Foundation inference model for temporal point processes}
\newacronym{desjepa}{DES-JEPA}{Joint-embedding predictive architecture for discrete event sequences}
\newacronym{despfn}{DES-PFN}{Discrete event sequence prior-fitted network}

\title{DES-JEPA: Learning a Shared Vocabulary for High-Dimensional Discrete Event Sequences}

\author{%
  Hugo Math \\
  BMW Group \\
  \texttt{hugo.math@bmwgroup.com} \\
  \And
  Maximilian Forstenh\"ausler \\
  BMW Group \\
  University of Glasgow \\
  \texttt{Maximilian.Forstenhaeusler@bmw.de} \\
  \AND
  Christos Anagnostopoulos \\
  University of Glasgow \\
  \texttt{Christos.Anagnostopoulos@glasgow.ac.uk}
}

\newenvironment{customitemize}{
  \begin{list}{\textbullet}{
    \setlength{\itemsep}{5pt}
    \setlength{\parsep}{0pt}
    \setlength{\topsep}{0pt}
    \setlength{\partopsep}{0pt}
    \setlength{\leftmargin}{10pt}
  }
}{
  \end{list}
}

\begin{document}

\maketitle

\begin{abstract}

\glslink{tfm}{Temporal foundation models} such as \glslink{chronos}{Chronos}\glsunset{chronos} and \glslink{timesfm}{TimesFM}\glsunset{timesfm} generalize across domains through a shared numerical observation space.\ In contrast, \glslink{des}{discrete event sequences (DESs)}\glsunset{des} including operational event logs or patient clinical-event trajectories have domain-specific event vocabularies whose labels lack a consistent correspondence across domains. Beyond this lack of consistency, the number of distinct events in real-world settings is high enough to challenge both classical \glslink{tpp}{temporal point process} modeling and current in-context foundation inference models, such as \acrshort{fimpp}. Critically, no current foundational method scales past a \emph{few dozen event types} for \glspl{des}, limiting their applicability to realistic, high-dimensional event streams.\ %
To address this, we propose two complementary modules: (i) \glslink{desjepa}{\textbf{DES-JEPA}}, a \glslink{jepa}{joint-embedding predictive architecture} that learns a shared discrete vocabulary from synthetic stochastic processes with unaligned event vocabularies; and (ii) \glslink{despfn}{\textbf{DES-PFN}}\glsunset{despfn}, a discrete event sequence prior-fitted network that reuses the learned vocabulary for in-context next-event prediction.\ We demonstrate that, without task-specific fine-tuning, this framework achieves competitive \textbf{zero-shot} performance on unseen real-world datasets and scales up to hundreds of distinct event types.
\end{abstract}
\section{Introduction and Related Work}\label{sec:introduction}
\glslink{des}{Discrete event sequences (DESs)}\glsunset{des} arise ubiquitously across domains, from vehicle diagnostic logs~\citep{math2024harnessingeventsensorydata} to \glslink{ehr}{electronic health records (EHRs)}\glsunset{ehr}~\citep{MedBERT, johnson2023mimic, pmlr-v219-labach23a}. Learning across these domains requires handling event vocabularies whose labels have no consistent meaning across datasets: the same event type may denote a machine error in one system and a clinical observation in another. 
Concurrent work on \glslink{tfm}{temporal foundation models (TFMs)}\glsunset{tfm} treats a sequence as a continuous-valued signal and pretrains a shared numerical value that lives on a comparable scale~\citep{ansari2024chronos, das2024timesfm}. However, categorical event identifiers provide no equivalent basis for such an aligned observation space. 

Prior work~\citep{ogata1978statistical, neuralhawkesprocess, transformerhawkeprocess, hawke_sparse_mutual} leverages \glslink{tpp}{temporal point processes (TPPs)}\glsunset{tpp} for modeling event types and their time of occurrence~\citep{hawke_process_and_app_2025}. Conventional methods therefore rely on fitting expensive statistical models (e.g., Hawkes processes \cite{ogata1978statistical}) that are known to be impractical at hundreds or thousands of distinct event types. Importantly, they must be fitted from scratch for every new batch of data. Berghaus et al.~\citep{berghaus2026incontext} address this with a foundation inference model \glslink{fimpp}{FIM-PP}\glsunset{fimpp}, enabling \emph{in-context} learning of \glspl{tpp} without fine-tuning.\ However, \gls{fimpp} supports at most 22 event types, restricting its direct application to datasets with larger event vocabularies and substantially limiting its scope as a foundation model. Therefore, extending current methodologies to more realistic real-world domains with hundreds of discrete event types remains an important open challenge. 

We hypothesize that the missing component is a representation that transfers across processes, not a better fit to any single one. Therefore, the correct invariant to learn across anchor-free \glslink{dgp}{data-generating processes (DGPs)}\glsunset{dgp} is therefore not the symbol itself, but alphabets of shared patterns that recur across unaligned processes. We gather this representation in a learned \emph{codebook}, which is reusable across downstream tasks such as zero-shot next-event prediction or long-horizon forecasting~\cite{shou2025unifiedflowmatchinglong}.

\tikzset{
    des panel/.style={line width=.5bp, rounded corners=3bp, dash pattern=on 5bp off 3bp},
}
\begin{figure}
    \centering
\resizebox{\textwidth}{!}{\begin{tikzpicture}[scale=0.25]

\pgfmathsetmacro{\anchordesjepax}{0}
\pgfmathsetmacro{\anchordesjepay}{0}

\pgfmathsetmacro{\anchordespfnx}{\anchordesjepax+27.5}
\pgfmathsetmacro{\anchordespfny}{0}

\pgfmathsetmacro{\anchoroutx}{\anchordesjepax+54}
\pgfmathsetmacro{\anchorouty}{0}

\draw[des panel, draw=desBorder, fill=myPurple!5] (\anchordesjepax,0) rectangle (\anchordesjepax+26,15);
\draw[des panel, draw=desBorder, fill=myBlue!5] (\anchordespfnx,0) rectangle (\anchordespfnx+25,15);
\draw[des panel, draw=desBorder, fill=myOrange!5] (\anchoroutx,0) rectangle (\anchoroutx+10,15); 

\node (heading_desjepa) at (3.5,14) {\scriptsize \textbf{(a) DES-JEPA}};
\node[gray] (description_desjepa) at (5.55,12.75) {\tiny Learn a shareable vocabulary};

\draw[line width=0.225mm] (4.5,1.75) -- (4.5, 11);
\draw[line width=0.225mm, dashed] (2.5,3) -- (4.5, 3);
\draw[line width=0.225mm] (4.5,3) -- (7.5, 3);
\draw[line width=0.225mm] (1.5,10) -- (4.5, 10);
\draw[line width=0.225mm, dashed] (4.5,10) -- (6, 10);

\foreach \p/\c in {2.0/teal,3.0/orange,3.75/violet}
    \draw[\c, fill=\c, thick] (\p,10) circle (5pt);

\foreach \p/\c in {5.35/teal,6.85/orange}
    \draw[\c, fill=white, thick] (\p,3) circle (5pt);
    
\node (query_text) at (1.7,9.25) {\tiny $x_{< t}$};
\node (future_window_text) at (2.25,2.25) {\tiny $x_{t:t+w}$};

\node[draw=myOrange, fill=myOrange!25, rounded corners=2pt, minimum height=0.45cm, minimum width=0.6cm] (stat_emb_box) at (8.5, 10) {};
\node (stat_emb_text1) at (8.45, 10+0.375) {\tiny Stat.};
\node (stat_emb_text2) at (8.55, 10-0.3) {\tiny Emb.};
\draw[-stealth] (6.25, 10) -- (stat_emb_box);

\node[draw=teal, fill=teal!10, rounded corners=2pt, minimum height=0.7cm, minimum width=0.8cm] (context_encoder_box) at (12.5, 10) {};
\node (context_encoder_text1) at (12.45, 10+0.8) {\tiny Student};
\node (context_encoder_text2) at (12.55, 10) {\tiny Encoder};
\node (context_encoder_text3) at (12.5, 10-0.85) {\tiny $f_{\theta}$};
\draw[-stealth] (stat_emb_box) -- (context_encoder_box);

\node[draw=myPurple!80!black, fill=myPurple!10, rounded corners=2pt, minimum height=0.5cm, minimum width=0.85cm] (pred_box) at (17, 10) {};
\node (pred_box_text1) at (17, 10+0.375) {\tiny Predictor};
\node (l_jepa_box_text) at (17, 10-0.45) {\tiny $g_{\theta}$};
\draw[-stealth] (context_encoder_box) -- (pred_box);

\node (z_hat_text) at (20.3, 10) {\tiny $\hat z$};
\draw[-stealth] (pred_box) -- (z_hat_text);

\node[draw=myBlue, dashed, minimum width=4.3cm, minimum height=0.85cm, rounded corners=2pt] (inference_box) at (17.1, 3) {};
\node[myBlue] (inference_text) at (23, 0.75) {\tiny \textbf{inference}};

\node[draw=myOrange, fill=myOrange!25, rounded corners=2pt, minimum height=0.45cm, minimum width=0.6cm] (stat_emb_box_fw) at (10, 3) {};
\node (stat_emb_fw_text1) at (9.95, 3+0.375) {\tiny Stat.};
\node (stat_emb_fw_text2) at (10.05, 3-0.3) {\tiny Emb.};
\draw[-stealth] (7.75, 3) -- (stat_emb_box_fw);

\node[draw=myBlue, fill=myBlue!10, rounded corners=2pt, minimum height=0.7cm, minimum width=0.8cm] (target_encoder_box) at (14, 3) {};
\node (target_encoder_text1) at (13.95, 3+0.8) {\tiny Teacher};
\node (target_encoder_text2) at (14.05, 3) {\tiny Encoder};
\node (target_encoder_text3) at (14, 3-0.85) {\tiny $f_{\bar\theta}$};
\draw[-stealth] (stat_emb_box_fw) -- (target_encoder_box);

\node[draw=red!50!black, fill=red!10, rounded corners=2pt, minimum height=0.4cm, minimum width=0.95cm] (quant_box) at (18.5, 3) {};
\node (quant_text) at (18.5, 3) {\tiny Quantizer};
\draw[-stealth] (target_encoder_box) -- (quant_box);

\node[draw=gray, fill=white, rounded corners=2pt, minimum height=0.8cm, minimum width=0.95cm] (codebook_box) at (23.5, 3) {};
\node (codebook_text1) at (23.5, 3+1) {\tiny Codebook};
\node (codebook_text2) at (23.5, 3) {\tiny $\mathcal{Z}$};

\foreach \p/\c in {22.25/JungleGreen, 23/myYellow, 23.75/Maroon, 24.5/Cyan}
    \draw[\c, fill=\c, thick] (\p, 3-1.2) rectangle ++(10pt, 10pt);

\node[] (e_text) at (16.2, 2.25) {\tiny $e$};
\node[] (c_text) at (21.1, 2.25) {\tiny $c$};
\draw[-stealth,] (target_encoder_box) -- (quant_box);
\draw[-stealth,] (21.6, 3) -- (quant_box);

\node[gray] (z_text) at (12.7, 6) {\tiny EMA};
\draw[-stealth, dashed, gray] (context_encoder_box) -- (12.5, 6.5) -- (14, 6.5) -- (target_encoder_box);
\node[draw=gray, fill=white, rounded corners=2pt, minimum height=0.4cm, minimum width=1cm] (l_jepa_box) at (18.5, 7.5) {};
\node (l_jepa_box_text) at (18.5, 7.5) {\tiny $\mathcal{L}_{\text{JEPA}}$};
\draw[-stealth] (z_hat_text) -- (21.5, 10) -- (21.5, 7.5) -- (l_jepa_box);
\draw[-stealth] (14.75, 4.4) -- (14.75, 7.5) -- (l_jepa_box);
\node (e_text2) at (15.25, 7) {\tiny $e$};
\draw[-stealth] (codebook_box) -- (23.5, 7.5) -- (l_jepa_box);
\node (c_text2) at (22, 7) {\tiny $c$};
\draw[-, gray] (17.7, 6.7) -- (17.7, 6.1);
\draw[-stealth, dashed, gray] (17.7, 4.5) -- (17.7, 3.85);
\node[gray] (sg_text1) at (17.7, 5.7) {\tiny stop};
\node[gray] (sg_text2) at (17.75, 4.95) {\tiny grad.};
\node (z_text) at (19.25, 5.2) {\tiny $z$};
\draw[-] (19.25, 3.85) -- (19.25, 4.75);
\draw[-stealth] (19.25, 5.7) -- (19.25, 6.7);

\node (heading_despfn) at (\anchordespfnx+3.2,14) {\scriptsize \textbf{(b) DES-PFN}};
\node[gray] (description_despfn) at (\anchordespfnx+4, 12.75) {\tiny In-Context Learning};
\draw[line width=0.225mm] (\anchordespfnx+7.5, 9.25) -- (\anchordespfnx+7.5, 12);
\draw[line width=0.15mm, gray] (\anchordespfnx+3, 11.5) -- (\anchordespfnx+8, 11.5);
\draw[line width=0.15mm, gray] (\anchordespfnx+3, 10.75) -- (\anchordespfnx+8, 10.75);
\draw[line width=0.2mm] (\anchordespfnx+3, 9.5) -- (\anchordespfnx+8, 9.5);

\foreach \p/\c in {3.5/violet,4.75/teal,5.85/orange, 6.5/violet}
        \draw[\c, fill=\c, thick] (\anchordespfnx+\p,11.5) circle (5pt);

\foreach \p/\c in {3.75/teal,4.5/orange,5.5/violet,6.9/teal}
        \draw[\c, fill=\c, thick] (\anchordespfnx+\p, 10.75) circle (5pt);

\foreach \p/\c in {3.6/orange,4.3/violet,6/teal,6.7/orange}
        \draw[\c, fill=\c, thick] (\anchordespfnx+\p, 9.5) circle (5pt);

\node (contex_text1) at (\anchordespfnx+1.5, 11.5) {\tiny $\mathcal{C}_1$};
\node (contex_text1) at (\anchordespfnx+1.5, 10.75) {\tiny $\mathcal{C}_2$};
\node (q_text1) at (\anchordespfnx+1.5, 9.5) {\tiny $q_{< t}$};

\node[draw=gray, fill=white, rounded corners=1pt, minimum width=1.65cm, minimum height=0.3cm] (codebook_pfn_box) at (\anchordespfnx+4.85, 7) {};
\node (codebook_pfn_box_text) at (\anchordespfnx+4.85, 7) {\tiny $\mathcal{Z}$};
\draw[draw=myBlue, fill=myBlue, rounded corners=0.5pt] (\anchordespfnx+1.3+6.3,7.3) rectangle (\anchordespfnx+2.1+6.3,7.9);
\draw[myBlue, line width=.5bp] (\anchordespfnx+1.45+6.3, 7.9) arc[start angle=180,end angle=0,x radius=0.25,y radius=.35];
\draw[-stealth, myBlue, line width=0.2mm] (inference_box) -- (26.75, 3) --  (26.75, 7) -- (codebook_pfn_box);
\node[draw=gray!30, fill=gray!5, rounded corners=2pt, minimum width=1.1cm, minimum height=0.65cm] (context_tokenized_box) at (\anchordespfnx+4.85, 3.75) {};
\foreach \xx/\kk/\c in {3/1/JungleGreen,4/3/Maroon,5/2/myYellow,6/1/JungleGreen}{
    \draw[rounded corners=1bp,line width=.6bp,draw=\c,
        fill=\c!20] (\anchordespfnx+\xx,4) rectangle ++(0.8,0.8);
    \node[\c] (token\kk) at (\anchordespfnx+\xx+0.4, 4+0.4) {\tiny \kk};
}
\foreach \xx/\kk/\c in {3/4/Cyan,4/2/myYellow,5/1/JungleGreen,6/3/Maroon}{
    \draw[rounded corners=1bp,line width=.6bp,draw=\c,
        fill=\c!20] (\anchordespfnx+\xx,2.7) rectangle ++(0.8,0.8);
    \node[\c] (token\kk) at (\anchordespfnx+\xx+0.4, 2.7+0.4) {\tiny \kk};
}
\foreach \xx/\kk/\c in {3/2/myYellow,4/3/Maroon,5/4/Cyan,6/1/JungleGreen}{
    \draw[rounded corners=1bp,line width=.6bp,draw=\c,
        fill=\c!20] (\anchordespfnx+\xx,1) rectangle ++(0.8,0.8);
    \node[\c] (token\kk) at (\anchordespfnx+\xx+0.4, 1+0.4) {\tiny \kk};
}
\node (contex_emb_text1) at (\anchordespfnx+1.5,4) {\tiny $\mathcal{C}_{z, 1}$};
\node (contex_emb_text1) at (\anchordespfnx+1.5,2.75) {\tiny $\mathcal{C}_{z, 2}$};
\node (q_emb_text1) at (\anchordespfnx+1.5,1) {\tiny $q_{z,< t}$};

\draw[-stealth] (\anchordespfnx+4.85, 9) -- (\anchordespfnx+4.85, 8);
\draw[-stealth] (\anchordespfnx+4.85, 6.2) -- (\anchordespfnx+4.85, 5.35);

\node[draw=myOrange, fill=myOrange!10, rounded corners=2pt, minimum height=0.7cm, minimum width=1.15cm] (causal_transformer_box_2) at (\anchordespfnx+14+0.4, 10+0.4) {};
\node[draw=myOrange, fill=myOrange!10, rounded corners=2pt, minimum height=0.7cm, minimum width=1.15cm] (causal_transformer_box_1) at (\anchordespfnx+14+0.2, 10+0.2) {};
\node[draw=myOrange, fill=myOrange!10, rounded corners=2pt, minimum height=0.7cm, minimum width=1.15cm] (causal_transformer_box) at (\anchordespfnx+14, 10) {};

\node (causal_transformer_text1) at (\anchordespfnx+14, 10+0.8) {\tiny Causal};
\node (causal_transformer_text2) at (\anchordespfnx+14, 10) {\tiny Transformer};
\node (causal_transformer_text3) at (\anchordespfnx+14, 10-0.85) {\tiny \textit{(temporal)}};
\node[draw=myPurple, fill=myPurple!10, rounded corners=2pt, minimum height=0.7cm, minimum width=1.1cm] (col_wise_attn_box_2) at (\anchordespfnx+14+0.4, 5+0.4) {};
\node[draw=myPurple, fill=myPurple!10, rounded corners=2pt, minimum height=0.7cm, minimum width=1.1cm] (col_wise_attn_box_2) at (\anchordespfnx+14+0.2, 5+0.2) {};
\node[draw=myPurple, fill=myPurple!10, rounded corners=2pt, minimum height=0.7cm, minimum width=1.1cm] (col_wise_attn_box) at (\anchordespfnx+14, 5) {};
\node (col_wise_attn_text1) at (\anchordespfnx+14, 5+0.8) {\tiny Col-wise};
\node (col_wise_attn_text2) at (\anchordespfnx+14, 5) {\tiny Attention};
\node (col_wise_attn_text3) at (\anchordespfnx+14, 5-0.85) {\tiny (\textit{in-context})};
\draw[-stealth, gray] (\anchordespfnx+7.2, 3.75) -- (\anchordespfnx+10, 3.75) -- (\anchordespfnx+10, 10.5) -- (\anchordespfnx+11.5, 10.5);
\draw[-stealth] (\anchordespfnx+7.2, 1.4) -- (\anchordespfnx+10.75, 1.4) -- (\anchordespfnx+10.75, 9.75) -- (\anchordespfnx+11.5, 9.75);
\draw[-stealth] (\anchordespfnx+14, 8.55) -- (\anchordespfnx+14, 6.85);
%
%
\node[draw=green!50!black, fill=green!10, rounded corners=2pt, minimum height=0.5cm, minimum width=1.1cm] (next_query_head_box) at (\anchordespfnx+20, 7) {};
\node (next_query_head_text1) at (\anchordespfnx+20, 7+.25) {\tiny Next-Query};
\node (next_query_head_text2) at (\anchordespfnx+20, 7-0.5) {\tiny Head};
\draw[-stealth] (\anchordespfnx+16.6, 5) -- (\anchordespfnx+20, 5) -- (\anchordespfnx+20, 5.95);
\node[draw=gray, fill=white, rounded corners=2pt, minimum height=0.3cm, minimum width=0.8cm] (softmax_box) at (\anchordespfnx+22.5, 10) {};
\node (softmax_text1) at (\anchordespfnx+22.5, 10) {\tiny softmax};
\draw[-stealth] (\anchordespfnx+20, 8) -- (\anchordespfnx+20, 10) -- (softmax_box);
\node[draw=gray, fill=white, rounded corners=2pt, minimum height=0.3cm, minimum width=1.4cm] (context_pairs_box) at (\anchordespfnx+20, 1.15) {};
\node (context_pairs_text) at (\anchordespfnx+20, 1.15) {\tiny Context Pairs};
\draw[-stealth, dashed, gray] (\anchordespfnx+8.2, 11) -- (\anchordespfnx+9.2, 11) -- (\anchordespfnx+9.2, 2.5) -- (\anchordespfnx+20, 2.5) -- (context_pairs_box);
\draw[-stealth, dashed, gray] (\anchordespfnx+8.25, 3.75) -- (\anchordespfnx+8.25, 1.15) -- (\anchordespfnx+17.15, 1.15);

\draw[-stealth, myBlue] (softmax_box) -- (\anchoroutx+1, 10);

\node (heading_out) at (\anchoroutx+4.7,14) {\scriptsize \textbf{(c) Forecasted Query}};
\node[gray] (description_despfn) at (\anchoroutx+4.1, 12.75) {\tiny Next-Event Prediction};

\node[black] (decode_text) at (\anchoroutx+5, 11.25) {\tiny $p_{\phi}(z_{t}\mid z_{< t}, \mathcal{C})$};
\draw[line width=0.225mm] (\anchoroutx+1.75, 2) -- (\anchoroutx+8.25, 2);
\draw[line width=0.15mm] (\anchoroutx+5, 1.25) -- (\anchoroutx+5, 2.5);

\foreach \p/\c in {2/violet,3/teal,4/orange}
    \draw[\c, fill=\c, thick] (\anchoroutx+\p+0.25, 2) circle (5pt);
\foreach \p/\c in {6/violet,7/teal,8/orange}
    \draw[\c, fill=white, thick] (\anchoroutx+\p-0.25, 2) circle (5pt);
\node[gray] (history_out) at (\anchoroutx+3.5, 3) {\tiny history};
\node[] (t_out) at (\anchoroutx+4.25, 1.15) {\tiny $t\text{-}\kern-0.1em1$};
\node[] (t_out) at (\anchoroutx+5.75, 1.15) {\tiny \(t\)};
\node[green!50!black] (forecast_out) at (\anchoroutx+6.75, 3) {\tiny forecast};
\draw[-stealth] (\anchoroutx+5.6, 6.85) -- (\anchoroutx+5.6, 6);
\node[draw=gray, fill=white, rounded corners=2pt, minimum height=0.4cm, minimum width=1cm] (kernel_decoder_box) at (\anchoroutx+5.6, 5.15) {};
\node (kernel_decoder_box_1) at (\anchoroutx+5.6, 5.55) {\tiny Kernel};
\node (kernel_decoder_box_2) at (\anchoroutx+5.6, 4.8) {\tiny Decoder};
\draw[-stealth] (\anchoroutx+5.6, 4.3) -- (\anchoroutx+5.6, 3.5);
\draw[-stealth, dashed, myBlue] (\anchordespfnx+23, 1.15) -- (\anchoroutx-0.75, 1.15) -- (\anchoroutx-0.75, 5.15) -- (kernel_decoder_box);


%

\foreach \xxx/\yyy/\l/\c in {2.5/1.1/1/JungleGreen!40, 3.5/2.3/2/JungleGreen, 4.5/1.5/3/JungleGreen!60}{
    \draw[\c, fill=\c, line width=0.15mm] (\anchoroutx+\xxx, 8) rectangle ++(0.6,\yyy);
    \node[\c] (label\l) at (\anchoroutx+\xxx+0.3, 7.35) {\tiny \l};
}
\node[gray] (label4) at (\anchoroutx+6, 8.35) {\tiny ...};
\node[gray] (label41) at (\anchoroutx+6, 7.35) {\tiny ...};
\draw[JungleGreen!20, fill=JungleGreen!20, line width=0.15mm] (\anchoroutx+6.75, 8) rectangle ++(0.6, 0.9);
    \node[JungleGreen!20] (label5) at (\anchoroutx+7.1, 7.35) {\tiny K};
\draw[line width=0.15mm] (\anchoroutx+2, 8) -- (\anchoroutx+8, 8);

\end{tikzpicture}
}
    \caption{\textbf{Pipeline overview.} \emph{(a):} an event sequence is tokenized into a shared codebook $\cZ$ via \textbf{DES-JEPA}. \textbf{DES-PFN} (b) learns in-context to predict the next code of a query $q_{<t}$. \emph{Right (c):} a sequence of interest is forecasted, the resulting code is then decoded into the event types.}
    \label{fig:method-overview}
\end{figure}
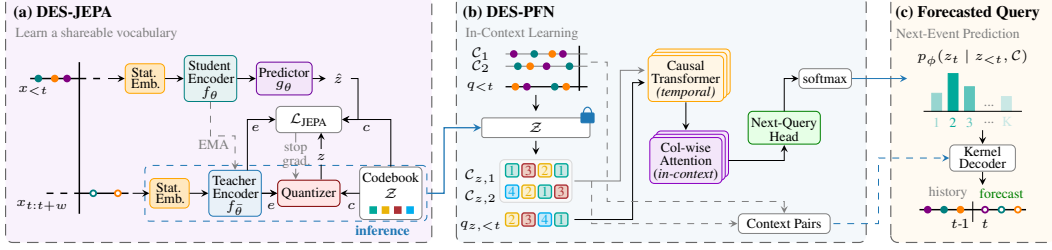


To address this gap, we introduce a \glslink{desjepa}{\textbf{j}oint-\textbf{e}mbedding \textbf{p}redictive \textbf{a}rchitecture for \textbf{d}iscrete \textbf{e}vent \textbf{s}equences (\textbf{DES-JEPA})}\glsunset{desjepa}, a \glslink{jepa}{JEPA}\glsunset{jepa}-based method~\citep{lecun2022path,assran2023ijepa,bardes2024vjepa} that learns a shared discrete vocabulary, i.e., \emph{codebook}. It consists of two transformer-based encoders: (i) a teacher encoder which quantizes future windows into latent codes, and (ii) a student encoder which maps historical observations to a latent representation. Through a \glslink{pfn}{prior-fitted network (PFN)}\glsunset{pfn}-style meta-pretraining (TabPFN, TabICLv2~\citep{muller2022pfn,hollmann2023tabpfn,hollmann2025tabpfnv2, qu2026tabiclv}) using synthetic data, the model learns to capture recurring dynamics across stochastic processes with unaligned vocabularies. Via the frozen teacher encoder and codebook embeddings, we tokenize any sequence of interest (\textit{query}). Then, a second model (\textbf{DES-PFN}) predicts the next-event on this \textit{query}, without fine-tuning, conditioned on a reference set of sequences (\textit{context}). Together, they support zero-shot next-event prediction through a shared latent vocabulary, extending amortized in-context inference to discrete representations of temporal dynamics. Our contributions are:
\begin{customitemize}
\item We introduce \textbf{\gls{desjepa}}, combining predictive latent learning, vector quantization, and meta-pretraining to learn a \textbf{shared codebook} from any discrete event sequences.
\item We introduce \textbf{\gls{despfn}}, which uses the shared codebook computed by the teacher encoder of \gls{desjepa}, to autoregressively predict unseen event types 
without task-specific fine-tuning. We evaluate it on zero-shot next-event prediction on various unseen real-world datasets ranging from 3 to 700 event types, placing it as the best performing model for event sequences.
\end{customitemize}

\section{Methodology}\label{sec:method}

\paragraph{Problem Statement.} 

Let $\cM$ denote a meta-distribution over a \gls{dgp}, for instance a high-order Markov chain \cite{koller2009probabilistic}. Each draw $m \sim \cM$ instantiates a process $\{X_t\}_{t \ge 0}$ over a finite, $m$-specific alphabet $\cX_m$, which changes identity with every drawn $m$. Thus, no embedding table indexed by a local event type \(x_i \in \mathcal{X}_m\) is shareable. Given any process, we aim to learn a shared representation that is invariant to event types $x_i$ that enables downstream tasks such as the modeling of $P(x_t|x_{<t})$ (Section \ref{sec:despfn}). 


\subsection{DES-JEPA: A Shared Discrete Codebook}\label{sec:desjepa}


%

Following \gls{jepa} \cite{lecun2022path, baek2025discretejepa}, \gls{desjepa} consists of two encoding paths: (i) an \textbf{encoder (student)} $f_\theta$ maps past observations to a representation $h = f_\theta(x_{<t})$, which are then mapped via a predictor $g_\theta$ to logits $\hat z = g_\theta(h)$ over the $K$ codes of $\cZ$, and (ii) a \textbf{\emph{bidirectional} encoder (teacher)} $f_{\bar\theta}$, a shallow prefix of $f_\theta$ which is only updated by an exponential moving average (EMA)~\citep{assran2023ijepa,bardes2024vjepa}, encodes the next raw events $x_{t:t+w}=(x_{t},\dots,x_{t+w})$ into $e = f_{\bar\theta}(x_{t:t+w}) \in \mathbb{R}^d$, quantized against $\cZ = \{c_1,\dots,c_K\}$ by nearest-neighbor lookup as,
\begin{equation}
z = k^*(x_{t:t+w}) = \arg\min_{k \in \{1,\dots,K\}} \lVert e - c_k \rVert_2^2, \quad \text{ where } w\geq1.
\label{eq:quantize}
\end{equation}
We keep $w=2$ fixed. The student never observes $x_{t:t+w}$ and predicts autoregressively $\hat z$ purely from $x_{< t}$, while the teacher is evaluated under stop-gradient $\sg[\cdot]$ and receives no gradient at all. This prevents both encoders from collapsing to a constant map \cite{assran2023ijepa}. 
For a single trajectory \(x_{<t}\) drawn from $m \sim \cM$, we train $f_\theta, g_\theta$ by minimizing, 
\begin{equation}
\mathcal{L}_{\text{JEPA}} = \mathbb{E}_{m \sim \cM}\ \mathbb{E}_{(x_{<t}, x_{t:t+w}) \sim m}\Big[\ \mathrm{CE}_\epsilon\big(g_\theta(h),\, z\big) \;+\; \big\lVert \sg[e] - c_k \big\rVert_2^2 \;+\; \beta \big\lVert e - \sg[c_k] \big\rVert_2^2\ \Big],
\label{eq:ltotal}
\end{equation}
at every position $t$. The main training signal is the cross-entropy loss, predicting the next latent code $\hat z$ from history $h$. A standard VQ-VAE codebook commitment terms~\citep{oord2017vqvae} ($\beta{=}0.25$) pulls $c_k$ and $e$ toward each other. Since $f_{\bar\theta}$ is trained only via EMA updates, the commitment term never actually back-propagates through $f_{\bar\theta}$. During inference, the future tokens $z_{t:t+w}$ are replaced by the embedding of the token $z_t$ to prevent any information leakage.


During training, sampling $m \; {\sim} \; \cM$ realizes amortized Bayesian PFN-style meta-training~\citep{muller2022pfn,hollmann2023tabpfn,hollmann2025tabpfnv2}: the codebook acts as a finite statistic for $m$'s identity, rather than representing $m$ directly (Appendix~\ref{app:mixture_of_dgp}). Naive VQ training is well known to collapse onto a small subset of indices. Thus, we actively monitor and enforce codebook utilization via a perplexity regularization and revive dead codes with a usage-EMA criterion. For additional details, please refer to 
Appendix~\ref{app:analysis}.

\subsection{DES-PFN: In-Context Autoregressive Next-Event Prediction}\label{sec:despfn}
%
Through the creation of codebook $\cZ$, we created a common encoding of \glspl{des}. This encoding is leveraged to pretrain a model on next-event prediction tasks in order to perform downstream zero-shot next-event prediction on unseen data. To do so, we leverage \emph{in-context learning} of event sequences \cite{berghaus2026incontext}, where for a sequence of interest \(q = ((x_0, t_0), \cdots, (x_t, t_t))\) (\textit{query}) given a \(\textit{context}\) \(\mathcal{C}=\{S_0, \cdots, S_C\}\), both drawn from the same \gls{dgp} \(m\), one performs the next-event prediction.



We freeze the codebook $\cZ$ and use it to tokenize each element in the raw sequence $q$ to form a sequence of event codes $q_z = (z_1, \dots, z_L)$ via the teacher path of Eq.~\eqref{eq:quantize}. We train \gls{despfn} on tokenized queries $q_z$ and context $\mathcal{C}_{z}$ generated from the same synthetic \glspl{dgp} meta-distributions $(\cM)$ used in the training of \gls{desjepa}. Given these inputs, \gls{despfn} predicts the next token $\hat q_{z,\,t}$ as, 
\begin{equation}\label{eq:despfn}
       \hat{q}_{z,\,t} = \arg\max p_\phi(q_{z,\,t} \mid q_{z,\,< t}, \cC_z),\,\,\,\text{ where }\,\,\,p_\phi = \text{softmax}\big(h_\phi(q_{z, < t}, \cC_z)\big).
\end{equation}

At \emph{inference}, the frozen \gls{desjepa} teacher maps the context $\cC$ and query history to discrete codes. \gls{despfn} then forecasts autoregressively in this shared code space, conditioning on the encoded inputs and previously predicted codes. This representation enables zero-shot transfer across \glspl{dgp} with disjoint event vocabularies, without updating either model. %
In the experiments, \gls{desjepa} and \gls{despfn} are pretrained on six \glspl{dgp} with a maximum codebook capacity of $K=768$ and contain 60 million and 40 million parameters, respectively. Architectural details are provided in Appendix~\ref{appendix:arch_details}.

%

\begin{figure}[t]
\centering
\includegraphics[width=\linewidth]{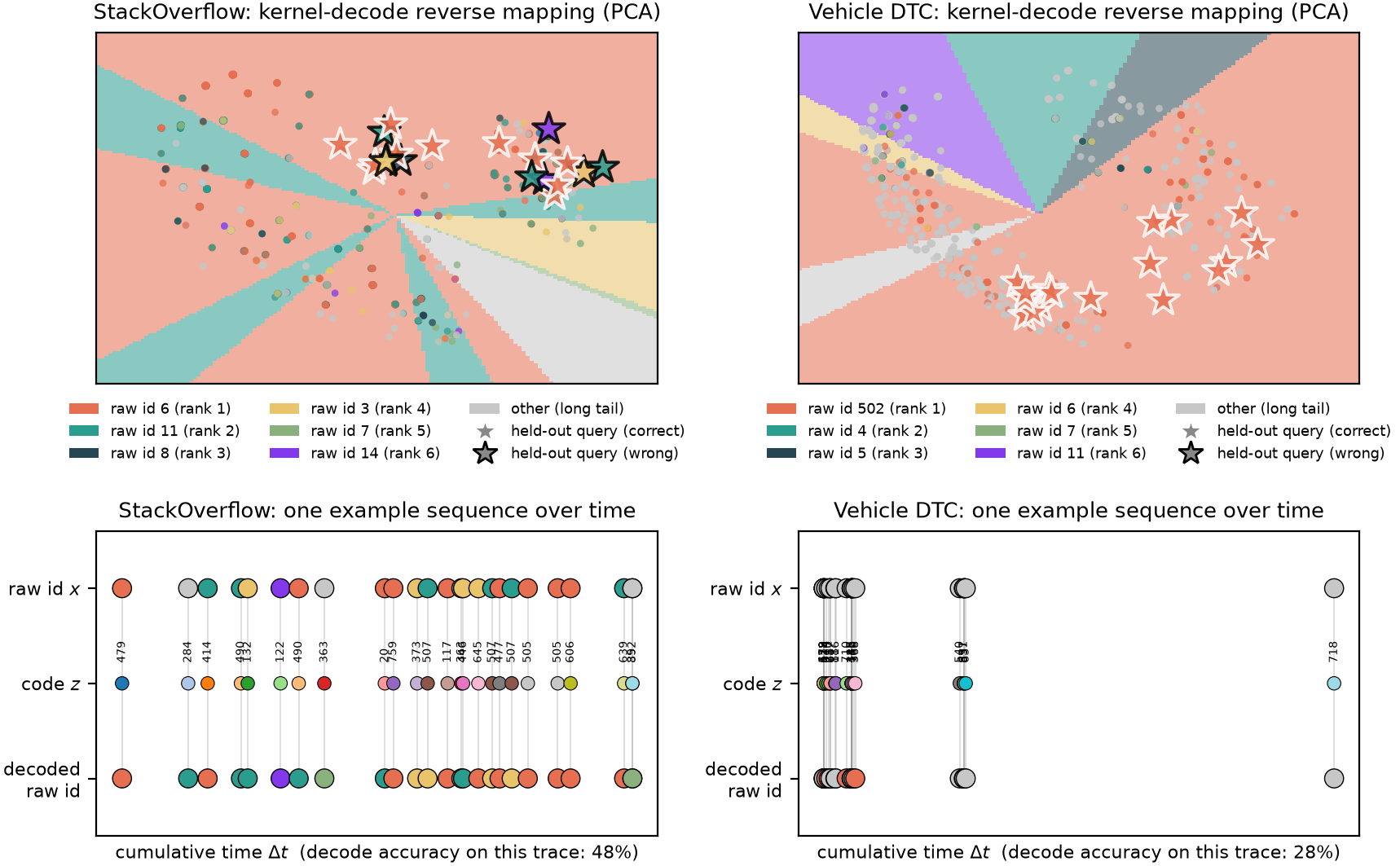}
\caption{\textbf{Top}: Kernel-decode inverse mapping on real StackOverflow and Vehicle DTC contexts, generated directly from the trained checkpoint (no illustrative numbers). \textbf{Bottom row}: the encode/decode round trip on ONE real sequence over cumulative time: raw ids, the codes DES-JEPA tokenizes them into, and the raw ids recovered by decoding those codes back. The color coding matches the top row, so a mismatch is directly visible as a color change.}
\label{fig:kernel-decode}
\end{figure}

\paragraph{Decoding Codes to Event Types.}
\gls{despfn} forecasts a tokenized code $\hat z=\hat q_{z,t}$, whereas the actually desired \textbf{event} prediction is required in its raw alphabet $\cX_m$. Since event labels are unaligned across processes and quantization is a many-to-one projection and therefore non-invertible, we estimate a context-specific token from the tokenized pairs $\{(x_i,z_i)\}_{i=1}^{C}$ in $\cC$. We use a Nadaraya-Watson estimator~\citep{nadaraya1964estimating,watson1964smooth} for candidate $v$,
\begin{equation}
\hat p(x=v\mid\hat z) = \frac{\sum_{i=1}^C \mathbb{1}[x_i=v]\, \kappa_\tau(e_{\hat z}, e_{z_i})}{\sum_{i=1}^C \kappa_\tau(e_{\hat z}, e_{z_i})},
\label{eq:kernel-decode}
\end{equation}
where $\kappa_\tau$ is a nonnegative cosine-based kernel with temperature $\tau$ calibrated per context and $v \in \cX_m$ a candidate event type. The predicted code $\hat z$ queries each context occurrence's code $z_i$, which casts a similarity-weighted vote for its identity $x_i$, allowing identities associated with nearby codes to be recovered without an exact match to $\hat z$ in $\cC$ (Fig.~\ref{fig:kernel-decode}).

\section{Experiments}\label{sec:experiments}
\label{sec:realworld-scale}
\paragraph{Baselines, Datasets \& Experimental Setup.}
We evaluate \textbf{\gls{despfn}} against FIM-PP~\citep{berghaus2026incontext} and three gradient-trained neural \gls{tpp} baselines, RMTPP~\citep{rrnembedding}, NHP~\citep{mei2017neuralhawkes}, and SAHP~\citep{selfatthawke}, for next-event prediction tasks. We benchmark the model performance on eight real event-type datasets: Twitter retweet cascades, StackOverflow badge awards, Taobao and Amazon user actions~\citep{xue2024easytpp}, HDFS and BGL system logs~\citep{he2020loghub,xu2009detecting,oliner2007supercomputers}, MIMIC-IV hospital lab-test sequences~\citep{johnson2023mimic} and a vehicle event logs~\citep{math2024harnessingeventsensorydata}. Every zero-/few-shot method, i.e., \gls{despfn}, FIM-PP and each neural baseline's \emph{few-shot} variant (refit from scratch on only the in-context set), sees the identical $n_{\mathrm{context}}{=}128$ example sequences. \gls{despfn} and FIM-PP receive no gradient updates. We report the validation top-1 accuracy for each \emph{fully-trained} neural baseline variant trained for 5 epochs, as a reference upper bound.\ 

\begin{table}[t]
\caption{\textbf{Real-world next-event-type accuracy} (Top-1, mean$\pm$std averaged over 5 runs) for $n_{\mathrm{context}}{=}128, n_{query}{=}20$ and $L{=}32$. Baselines are reported in two regimes: \emph{few-shot} (refit from scratch on the same $128$ context sequences) and \emph{full} (fully-trained). For HDFS/BGL/MIMIC-IV, since no real train/test split is available, the `full' column instead reports a $^\S$-marked \emph{local train/test split} (trained once on every earlier sequence actually available, evaluated on the same held-out query set); `few-shot' stays ``--'' for these three (not run). FIM-PP is capped at $|\mathcal{X}| \le 22$ marks (``--''). Bold/underline mark the best/second-best score per row. Vehicle is reported at two vocabulary caps ($|\mathcal{X}|{=}500$ and $|\mathcal{X}|{=}700$), $\dagger$ marks a method that genuinely exceeded the $600\mathrm{s}$ time budget.}
\label{tab:realworld-scale}
\centering
\small
\resizebox{\textwidth}{!}{%
\begin{tabular}{lccccccccc}
\toprule
& & \multicolumn{2}{c}{RMTPP~\citep{rrnembedding}} & \multicolumn{2}{c}{NHP~\citep{mei2017neuralhawkes}} & \multicolumn{2}{c}{SAHP~\citep{selfatthawke}} & & \\
\cmidrule(lr){3-4}\cmidrule(lr){5-6}\cmidrule(lr){7-8}
Dataset & $|\mathcal{X}|$ & few-shot & full & few-shot & full & few-shot & full & FIM-PP~\citep{berghaus2026incontext} & \gls{despfn} (ours) \\
\midrule
retweet~\citep{xue2024easytpp} & 3 & 0.464$\pm$0.021 & \underline{0.533} & 0.235$\pm$0.011 & \textbf{0.591} & 0.531$\pm$0.020 & 0.471 & 0.462$\pm$0.022 & 0.520$\pm$0.023 \\
stackoverflow~\citep{xue2024easytpp} & 22 & 0.009$\pm$0.003 & 0.411 & 0.007$\pm$0.002 & \underline{0.420} & 0.413$\pm$0.035 & \underline{0.420} & \textbf{0.441$\pm$0.032} & 0.356$\pm$0.030 \\
taobao~\citep{xue2024easytpp} & 17 & 0.227$\pm$0.049 & \textbf{0.559} & 0.490$\pm$0.049 & \underline{0.522} & 0.433$\pm$0.041 & 0.465 & 0.516$\pm$0.040 & 0.420$\pm$0.057 \\
amazon~\citep{xue2024easytpp} & 16 & 0.151$\pm$0.028 & \underline{0.299} & 0.089$\pm$0.006 & \textbf{0.319} & 0.193$\pm$0.031 & 0.254 & 0.207$\pm$0.041 & 0.241$\pm$0.035 \\
HDFS~\citep{he2020loghub,xu2009detecting} & 14 & -- & 0.074$^\S$ & -- & 0.063$^\S$ & -- & 0.103$^\S$ & \textbf{0.266$\pm$0.018} & \underline{0.200$\pm$0.028} \\
BGL~\citep{he2020loghub,oliner2007supercomputers} & 120 & -- & 0.000$^\S$ & -- & 0.000$^\S$ & -- & \underline{0.063$^\S$} & -- & \textbf{0.136$\pm$0.020} \\
MIMIC-IV~\citep{johnson2023mimic} & 441 & -- & 0.000$^\S$ & -- & \underline{0.004$^\S$} & -- & 0.002$^\S$ & -- & \textbf{0.029$\pm$0.010} \\
Vehicle~\citep{math2024harnessingeventsensorydata} & $500$ & -- & $\dagger$ & -- & $\dagger$ & -- & $\dagger$ & -- & \textbf{0.162$\pm$0.014} \\
Vehicle~\citep{math2024harnessingeventsensorydata} & $700$ & -- & $\dagger$ & -- & $\dagger$ & -- & $\dagger$ & -- & \textbf{0.094$\pm$0.009} \\
\bottomrule
\end{tabular}}
\end{table}
\paragraph{Results.} Table~\ref{tab:realworld-scale} reports \gls{despfn}'s zero-shot (no-gradient-update) next-event prediction accuracy across eight real datasets spanning three to seven hundred event types. 
Fully-trained specialists gradient-trained on each dataset's entire training split still outperform \gls{desjepa} on the four lower-cardinality \texttt{easytpp} datasets. However, this specialism does not generalize well, since each needs a real held-out test split and per-dataset retraining from scratch to be accurate. In low cardinality, FIM-PP, which shares \gls{despfn}'s zero-shot, in-context inference setting has competitive performance beating \gls{despfn} in three out of five datasest. 
\emph{Importantly}, its fixed $22$-event type architecture precludes evaluation on BGL, MIMIC-IV, and Vehicle. Using the single frozen codebook, \gls{despfn} supports larger event vocabularies and is the only method with reported results on these datasets; all gradient-trained baselines either lack a usable data split or exceed the prescribed training-time budget. \gls{despfn}'s  weakest result is on MIMIC-IV: 
we reason that the shared blood-draw times tampes across $95\%$ of consecutive lab events may leave \gls{despfn} little time information to exploit.

\section{Conclusion}\label{sec:conclusion}

We introduced a framework to model high-dimensional event sequences and enable zero-shot next-event prediction. The first model, \textbf{\gls{desjepa}} pretrains a single, shared codebook over a meta-distribution of synthetic \glspl{dgp}, giving arbitrary, anchor-free event sequences that share no vocabulary or numeric scale, a common representation. By applying this frozen codebook, \textbf{\gls{despfn}} is able to perform zero-shot next-event prediction on eight real-world dataset spanning across three orders of magnitude in cardinality, from $K{=}3$ to $K{=}700$. This approach remains competitive at low cardinality, and remains the only reliable method producing a non-trivial number once cardinality grows to hundreds of event types. Two limitations motivate future work: (i) the codebook capacity is fixed during pretraining, and currently lacks rate-distortion experiments (ii) evaluation is restricted to one-step-ahead prediction. Long-horizon autoregressive rollouts remain untested, particularly regarding error accumulation and the ability to represent multiple plausible futures beyond a single point forecast.

\bibliography{neurips_2026}

\appendix
\section*{Appendix}\label{sec:app:appendix}

\section{Limitations.} 

Two limitations point to natural extensions. The codebook size is fixed at pretraining time rather than adapted to a downstream dataset's true cardinality. Moreoever, every result in this paper is one-step-ahead: DES-PFN has not been evaluated on genuine long-horizon, multi-step rollout, where a single point forecast poorly represents the many plausible futures a real event stream can take and small early errors can compound over many steps. We see these last two limitations as connected: the natural next step is to replace the single-code predictor with a multi-head decoder trained under a winner-takes-all, multiple-choice-learning objective~\citep{cortes2025winnertakesall}, letting DES-JEPA and DES-PFN output several plausible future codes per query instead of one point forecast, precisely the mechanism long-horizon rollout needs to stay calibrated as uncertainty compounds step by step.

\section{Architecture Detail}
\label{appendix:arch_details}

Both DES-JEPA and DES-PFN were pretrained on a single NVIDIA RTX 2080 Ti, approximately one day each. All experimental results were measured at inference time on an Apple M2 Pro CPU, with no GPU acceleration.

\subsection{DES-JEPA}

\subsubsection{Embeddings}
Every event is embedded from purely statistics of its own occurrence history: self-information/surprisal, running (global and local-window) frequency, whether the type has been seen before, and recency and dispersion of its own inter-arrival gaps, together with event-count (``step'') analogues of the timing features that need no real timestamp. None of these is computed from a learned lookup table keyed on the raw event id, since a raw id carries no stable meaning across draws of $m\sim\cM$ (Section~\ref{sec:method}): therefore event identity is represented purely through these statistics. We attach no additional per-event covariate beyond type and time (no ``marks'' in the temporal-point-process sense) to any data-generating process. 

\begin{figure}[!h]
    \centering
    \includegraphics[width=0.9\linewidth]{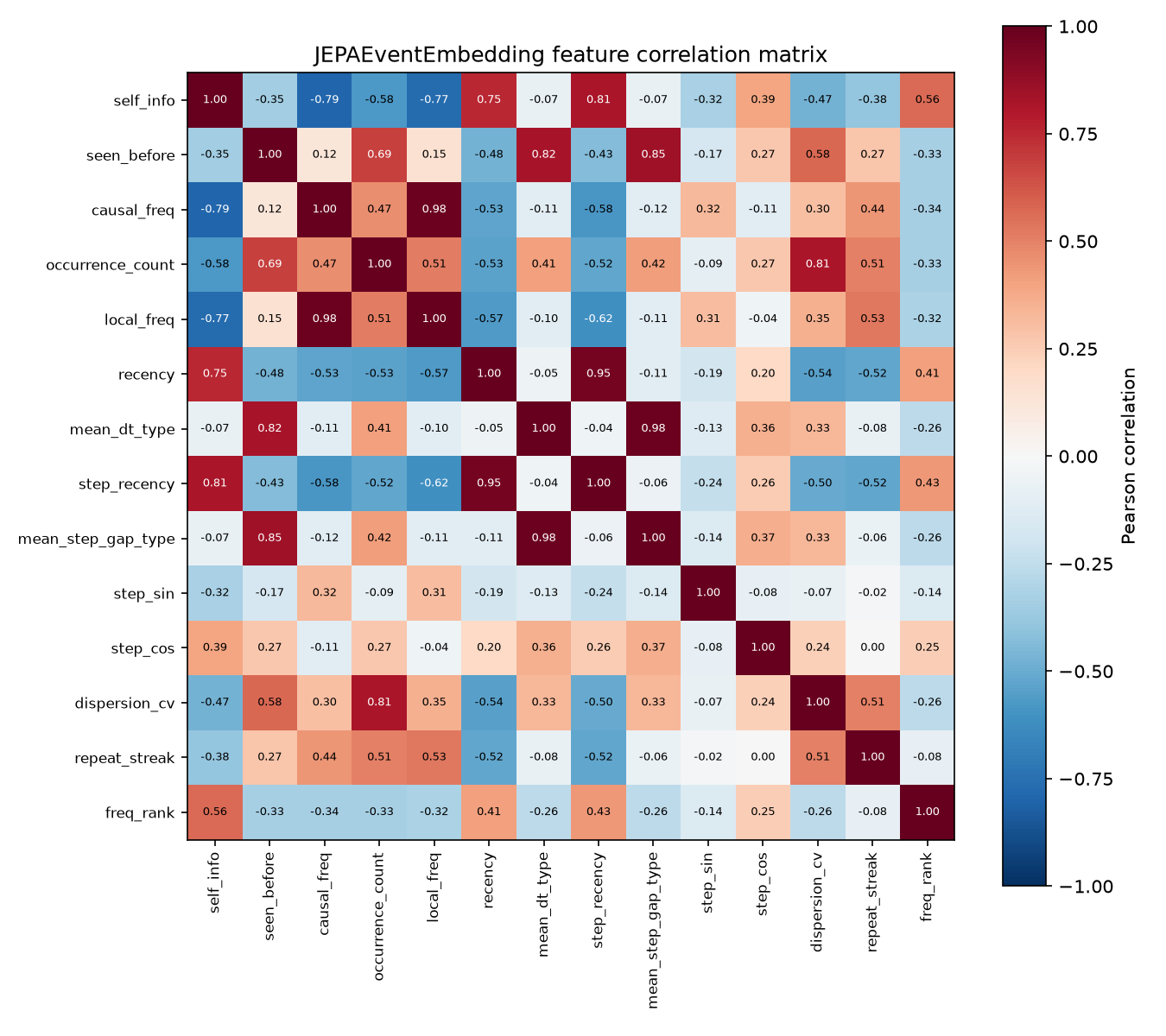}
    \caption{Pairwise Pearson correlation across the 14 causal statistical features feeding the DES-JEPA event embedding, computed on a mixed-DGP synthetic batch.}
    \label{fig:feature-corr}
\end{figure}

Fig.~\ref{fig:feature-corr} shows the pairwise correlation of these features on a representative synthetic batch: highly correlated pairs (e.g.\ \texttt{causal\_freq}/\texttt{local\_freq}, \texttt{mean\_dt\_type}/\texttt{mean\_step\_gap\_type}) are deliberately kept rather than pruned, since their real-valued/step-valued counterparts remain the only signal available whenever timestamps are missing or unreliable.

\subsubsection{Transformer Block}

\paragraph{Student Encoder.} A causal Transformer over the whole past, with sliding-window attention (window $w{=}32$: token $t$ attends only to the $32$ preceding tokens) and RoPE~\citep{Roformer} position encoding.

\paragraph{Teacher Encoder.} A shallow, \emph{bidirectional} prefix of the student stack ($2$ of the student's $6$ layers in the checkpoint used throughout this paper), updated only by an EMA of the student's own weights ($\tau{=}0.999$) (Section~\ref{sec:desjepa}): it encodes the next $W{=}2$ raw events as one coherent unit rather than causally within that short window.

\paragraph{Vector Quantizer.} The teacher's continuous output is matched to the nearest of $K$ codebook vectors Eq.~\eqref{eq:quantize} and trained with the combined cross-entropy/VQ-VAE objective of Eq.~\eqref{eq:ltotal}. Appendix~\ref{app:analysis} details the perplexity regularizer and dead-code revival mechanism that keep this codebook from collapsing into a dozen of codes.

\subsection{DES-PFN}

\subsubsection{Embeddings}
DES-PFN's event-code embedding table is the frozen DES-JEPA codebook itself (Section~\ref{sec:despfn}), followed by a learned linear projection whenever the two models' hidden sizes differ. Inter-arrival time uses a Chronos-style~\citep{ansari2024chronos} discretized embedding: $\log(1+\Delta t/\text{scale})$ is binned into a fixed vocabulary and looked up, turning next-time prediction into classification rather than continuous regression. An optional duplicate-count embedding additionally marks immediate event repeats.

\subsubsection{Transformer Block}
DES-PFN alternates two attention axes, at every layer, over the (sequence, time, hidden) grid formed by the reference context and the query.

\paragraph{Causal Attention.} A standard causal Transformer block over the temporal axis, applied independently to every sequence (context and query alike), with RoPE and a sliding-window band ($32$ in the checkpoint used throughout this paper).

\paragraph{Col-Wise Attention.} Following the column-then-row, induced-attention design recently used by tabular foundation models such as TabICLv2~\citep{qu2026tabiclv} and the inducing-point mechanism of Set Transformer~\citep{lee2019settransformer}, cross-sequence attention pools information through a small, fixed set of $n_\text{inducing}{=}32$ learnable vectors rather than full $O(n^2)$ attention across reference sequences. 

\section{Pretraining}
\label{app:mixture_of_dgp}

\subsection{Prior-Fitted Network and Synthetic Data Generation}
Both stages are pretrained exclusively on synthetic data drawn from a meta-distribution $\cM$ over six data-generating process (DGP) families, each implemented in \texttt{synthetic\_data/}. \emph{No real-world data is used at either pretraining stage}; every real-world result in Section~\ref{sec:realworld-scale} is zero-shot transfer. Every family is instantiated fresh per task (all parameters below redrawn) and draws its own active event-type count $|\cX|$ (the curriculum's $n_\text{active}$, denoted this way to avoid clashing with the codebook size $K$) \emph{uniformly} from a shared range, detailed at the end of this section.

\paragraph{Markov ($r$-th order chain).} Order $r\sim\mathrm{Unif}\{1,\dots,5\}$ is drawn once per task -- not fixed at $1$: the next type follows
\begin{equation}
P(X_t = k \mid X_{t-r:t-1}) = \mathrm{softmax}_k\Big(b_k + \textstyle\sum_{\ell=1}^{r} e^{-\ell}\, W[X_{t-\ell}, k] + f_\text{NL}(X_{t-r:t-1})_k\Big),
\end{equation}
with bias $b\in\mathbb{R}^{|\cX|}$, sparse mixed-sign interaction matrix $W\in\mathbb{R}^{|\cX|\times|\cX|}$ (density $1{-}\text{sparsity}$, $\text{sparsity}\sim U(0.3,0.8)$; $W_{j,k}{>}0$ excitatory, $W_{j,k}{<}0$ inhibitory), and $e^{-\ell}$ an exponential recency decay over the $r$ lags. With probability $0.5$ per task, a two-layer MLP term $f_\text{NL}$ (random fixed embedding, one hidden layer, a randomly chosen activation) adds non-additive lag interactions; otherwise $f_\text{NL}{\equiv}0$. Holding times are Weibull with a history-modulated scale: $\Delta t \mid X_t{=}k \sim \mathrm{Weibull}(\text{shape}_k, \text{scale}_t)$, $\text{scale}_t = \mathrm{softplus}\big(s_k + \textstyle\sum_{\ell=1}^{r} e^{-\ell/2} w[X_{t-\ell}]\big) + \epsilon$, $\text{shape}_k = \mathrm{softplus}(g_k) + 0.1$, where $s_k, w, g_k$ are fixed-at-construction (not trained) per-task parameters; $w$ depends only on the preceding types, not on the candidate $k$.

\paragraph{Hawkes (exact Ogata~\citep{ogata1978statistical} thinning).} A multivariate self-exciting process with conditional intensity
\begin{equation}
\lambda_k(t) = \mu_k + \sum_{j=1}^{|\cX|} \sum_{t_i^j < t} A_{j,k}\, \kappa(t - t_i^j),
\end{equation}
baseline $\mu_k \sim \mathrm{LogUniform}(0.05, 0.5)$; excitation $A\in\mathbb{R}_{\ge0}^{|\cX|\times|\cX|}$ built sparse (each row excites $1$--$3$ other types, sparsity$\sim U(0.2,0.5)$, weights $\sim\mathrm{Exp}(1)$) plus a diagonal $A_{k,k} = (\sum_j A_{k,j})\,u_k$, $u_k\sim U(0.2,0.5)$, globally rescaled so the spectral radius $\rho(A)\sim U(0.7,0.9)$ (stationarity, $\rho(A){<}1$). The kernel $\kappa$ is, with probability $0.5$ each: exponential $\kappa(\Delta)=\beta e^{-\beta\Delta}$, $\beta\sim U(1,5)$ (short-range, Markovian); or power-law/Omori--Utsu $\kappa(\Delta)=\alpha c^\alpha(\Delta+c)^{-(1+\alpha)}$, $\alpha\sim U(0.3,1.2)$, $c\sim U(0.05,0.5)$ (long-range, non-Markovian).

\paragraph{Neural Autoregressive Models} A randomly-initialized, never-trained LSTM: $h_t = \mathrm{LSTM}(e(X_t), h_{t-1})$, $X_{t+1}\sim\mathrm{Categorical}\big(\mathrm{softmax}(W_\text{tok}h_t)\big)$, $\Delta t_{t+1} \sim \mathrm{Weibull}\big(\mathrm{softplus}(w_\text{shape}^\top h_t),\, \mathrm{softplus}(w_\text{scale}^\top h_t)\big)$, where token and timing heads read from the SAME $h_t$ (fully joint dynamics) and $e$ is a random embedding table. Hidden size $\sim U\{16,\dots,64\}$, depth $\sim U\{1,2,3\}$ layers, embedding dimension $\sim U\{8,\dots,32\}$, all weights i.i.d.\ $\mathcal{N}(0,\sigma^2)$, $\sigma\sim U(0.5,2.0)$ -- freshly redrawn every task.

\paragraph{Hidden Markov Model.} $S\sim U\{3,\dots,7\}$ latent states, $z_{t+1}\mid z_t\sim\mathrm{Categorical}(T_{z_t,:})$; $T$ is sticky, with the \emph{escape} probability (not the sticky value itself) drawn log-uniformly, $1{-}p_\text{stick}\sim\mathrm{LogUniform}(0.01,0.3)$ (i.e.\ $p_\text{stick}\in[0.7,0.99]$) -- avoiding oversampling short dwell times, since dwell $=1/(1{-}p_\text{stick})$ is highly nonlinear in $p_\text{stick}$. Emission $X_t\mid z_t\sim\mathrm{Categorical}(E_{z_t,:})$, $E_{k,:}\sim\mathrm{Dirichlet}(\alpha\mathbf{1}_{|\cX|})$, $\alpha\sim U(0.2,0.5)$ (small $\alpha$ -- each state emits a small, peaked subset of types); timing $\Delta t_t\mid z_t\sim\mathrm{Weibull}(\text{shape}_{z_t},\text{scale}_{z_t})$, $\text{shape}_k\sim U(0.8,2.5)$, $\text{scale}_k\sim\mathrm{LogUniform}(0.3,3)$ -- the same latent state drives both what happens and when.

\paragraph{Burst.} An i.i.d.\ renewal timing process, $\Delta t_t\sim\mathrm{Weibull}(\text{shape},\text{scale})$, $\text{shape}\sim U(0.15,0.7){<}1$ (decreasing hazard, genuinely bursty), $\text{scale}\sim U(0.2,1.0)$, both fixed for the whole task and independent of type/history -- combined with a first-order sticky-categorical type chain: $X_t = X_{t-1}$ with probability $p_\text{stick}\sim U(0,0.9)$, else $X_t\sim\mathrm{Categorical}(p)$ resampled, $p\sim\mathrm{Dirichlet}(\alpha\mathbf{1}_{|\cX|})$, $\alpha\sim U(0.3,1.5)$. Models verbatim repeat-heavy streams (keep-alives, retries, repeated clicks).

\paragraph{Template.} A shared directed graph $\mathrm{next}: \{1,\dots,|\cX|\} \rightharpoonup \{1,\dots,|\cX|\}$ (at most one outgoing edge per node) is built from short chains, length $\sim U\{3,\dots,8\}$; a \emph{latent} walk deterministically follows $z_{t+1}=\mathrm{next}(z_t)$ whenever defined, otherwise restarts at a start node chosen $\propto$ per-template weights $\sim\mathrm{Dirichlet}(\alpha\mathbf{1})$, $\alpha\sim U(3,8)$. The \emph{emitted}, noisy type mixes a point mass on the latent node with a uniform background,
\begin{equation}
P(X_t = v \mid z_t) = (1-\epsilon)\,\mathbb{1}[v = z_t] + \epsilon / |\cX|, \qquad \epsilon \sim U(0, 0.1),
\end{equation}
so every modeled edge $(j, \mathrm{next}(j))$ has expected causal strength $1{-}\epsilon$ -- this is what gives event-type \emph{pairs} a near-deterministic causal coupling that Hawkes-style self-excitation (which only boosts future \emph{rate}, never identity) cannot express. Timing is $\Delta t_t \sim \mathrm{Weibull}(1, s_t)$ (i.e.\ Exponential), with scale $s_t = s_\text{slow}\sim U(0.5,3)$ right after a template restart and $s_t = s_\text{fast}\sim U(0.05,0.3)$ while continuing along an established edge.

\paragraph{Sampling Parameters During Pretraining.}
DES-JEPA (Stage 1) samples a single trajectory per task, from all six DGPs with \textsc{template} and \textsc{hawkes} up-weighted to $30\%$ and $15\%$ probability respectively (the remaining four share the rest evenly, $13.75\%$ each), a fixed sequence length of $128$, and event-type cardinality $n_\text{active}$ drawn log-uniformly from $[3, 1024]$ per task. DES-PFN (Stage 2) instead samples both a reference context $\cC$ and a query sequence per task, following a TabICLv2-style~\citep{qu2026tabiclv} curriculum that couples context size and sequence length under a fixed per-batch token \emph{budget} ($n_\text{context}\times\text{length}\approx\text{const}$): $n_\text{context}\in[16,256]$ and sequence length $\in[16,128]$ vary anti-correlated by construction, so a batch of many short reference sequences and a batch of few long ones cost roughly the same compute; $n_\text{active}\in[3,1024]$ as in Stage 1.

\paragraph{Tokenizing an observed sequence at inference (the last $w-1$ positions).} During \emph{training}, the teacher only ever supervises positions with a genuine, fully-observed future window: for a length-$L$ training sequence, positions $0,\dots,L{-}W{-}1$ each have a real $x_{t:t+W}$ target, and the remaining $W$ tail positions are simply dropped from the loss. At \emph{inference}, however, \gls{despfn} needs one code per position for the entire query, including its most recent, currently-last observed event, but that position's own future ($x_{t+1}$) is exactly what is being forecast and does not exist yet. The tokenization made by DES-JEPA resolves this by computing genuine windows for every position that has one (positions $0,\dots,L{-}W$) and, for the remaining $W{-}1$ tail position(s) only, substituting a degenerate window built by repeating the last observed embedding $W$ times (\texttt{emb[:, -1:].expand(...)}) in place of the unavailable future. With $W{=}2$ (the value used throughout this paper) this affects only a single position per tokenized sequence: the code assigned to the most recently observed event is computed from this repeated-embedding fallback rather than a genuine future window, for every context sequence and for the query alike. This is a real, deliberate train/inference asymmetry (never exercised during training, since the tail positions are excluded from the loss there) rather than a bug, but it does mean the single most recent position's code is systematically less informative than every earlier position's.

\section{Additional Analysis}
\label{app:analysis}

\paragraph{Codebook Perplexity.} A universal codebook shared across heterogeneous DGPs is only useful if its capacity is actually exploited; naive VQ training is well known to collapse onto a small subset of indices (e.g., DES-JEPA using 10 codes at most instead of the $K$ available ones). We track the empirical code-usage distribution over a batch of $B$ target windows, $\hat p_k = \frac{1}{B}\sum_{i=1}^B \mathbb{1}[k^*(y_i) = k]$, and define

\begin{figure}[!h]
    \centering
    \includegraphics[width=0.95\linewidth]{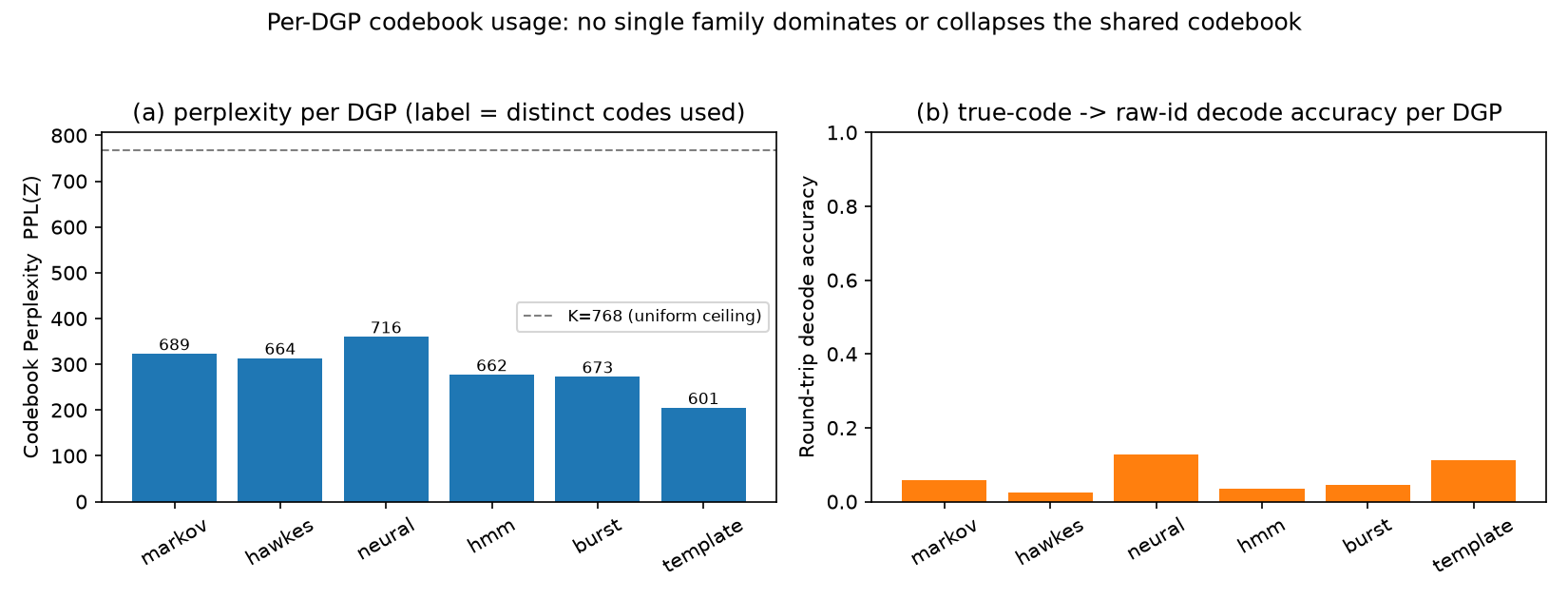}
    \caption{Per-DGP codebook Perplexity (left) and round-trip true-code-to-raw-id decode accuracy (right), measured directly from the trained checkpoint used throughout this paper (no retraining), $800$ fresh synthetic tasks per DGP family, $n_{\mathrm{context}}{=}32$.}
    \label{fig:dgp-codebook-stats}
\end{figure}

\begin{definition}[Codebook Perplexity]
\begin{equation}
\mathrm{PPL}(\cZ) = \exp\!\left(-\sum_{k=1}^K \hat p_k \log \hat p_k\right).
\label{eq:ppl}
\end{equation}
\end{definition}
$\mathrm{PPL}(\cZ) \in [1, K]$, with $\mathrm{PPL} = K$ under uniform code usage and $\mathrm{PPL} \to 1$ under index collapse onto a single code. We monitor $\mathrm{PPL}$ per-DGP-family and aggregated across $\cM$: a universal codebook whose perplexity remains high \emph{within} every family but whose per-family code supports remain \emph{distinguishable} across families is precisely the signature of a shared dictionary that is neither collapsed nor merely memorizing family identity. Aggregated across the full training mixture, $K{=}768$ codes give a final perplexity of $\approx 500$. Fig.~\ref{fig:dgp-codebook-stats} breaks this down per DGP family, measured directly from the trained checkpoint on $800$ fresh synthetic tasks per family ($n_{\mathrm{active}}\in[3,300]$): perplexity is high and comparable across all six families ($601$--$716$ of the $768$ available codes actually used, no family collapsing onto a handful or hogging most of the codebook), supporting the "shared, non-collapsed dictionary" reading above. The decode-accuracy panel (true code $\to$ raw id via the same per-context majority-vote decoder used for real-world data, Section~\ref{sec:despfn}) is uniformly modest ($2.5$--$12.7\%$) across every family: with up to $300$ active raw ids sharing only $768$ codes total across six families. Exact code identity alone cannot always pin down a unique raw id (the same codebook-capacity/collision cost documented for real-world high-cardinality datasets in the experiment section. An high perplexity number indicates the codebook is being used broadly and non-degenerately, not that every individual code is uniquely decodable.

\paragraph{Dead-code revival.} A perplexity regularizer only discourages collapse in aggregate; it does not repair codes that have already drifted out of use. Following the EMA-cluster-size restarts of VQ-VAE-2~\citep{razavi2019vqvae2} and its scaled-up variant in Jukebox~\citep{dhariwal2020jukebox}, we track an exponential moving average of each code's usage share,
\begin{equation}
\bar p_k^{(t)} \leftarrow \mu\, \bar p_k^{(t-1)} + (1-\mu)\, \hat p_k^{(t)}, \qquad \bar p_k^{(0)} = \tfrac{1}{K},
\label{eq:usageema}
\end{equation}
and flag code $k$ as \emph{underused} whenever $\bar p_k^{(t)}$ falls below a fraction $\vartheta$ of its fair share $1/K$. A per-code counter $d_k$ accumulates consecutive underused steps and resets to zero the moment $k$ recovers.
The code $k$ becomes reset-eligible once $d_k \ge T_{\mathrm{pat}}$. Unlike a hard, per-step presence test, $\bar p_k^{(t)}$ is barely perturbed by a single rare assignment, so a code that only survives on stray, noise-level picks is still correctly identified as dead. All simultaneously reset-eligible codes are re-seeded in one batched event, from encoder outputs $z$ of the current batch, once the reset-eligible fraction exceeds a threshold $\rho$,
\begin{equation}
\frac{1}{K}\sum_{k=1}^K \mathbb{1}[d_k \ge T_{\mathrm{pat}}] \;\ge\; \rho .
\label{eq:resetgate}
\end{equation}
Relative to the per-code restarts of prior work, our two departures are: (i) a batched trigger that rescues a whole backlog of dead codes in a single event rather than continuously reviving them one at a time, and (ii) a usage-\emph{share}-relative threshold, rather than an absolute cluster-size count, which keeps the criterion scale-invariant as $B \times L$ grows far beyond $K$.

\section{Related Work}\label{sec:related-work}
\paragraph{Transfer across temporal processes.} Continuous \glspl{tfm} such as Chronos, TimesFM, and UniTok transfer forecasts through a shared numerical observation space \citep{ansari2024chronos,das2024timesfm,zhang2026unitok}. \glspl{pfn}, \glspl{fim}, MetaTPP and LaT-PFN instead amortize inference across tasks or stochastic processes, but transfer continuous forecasts, intensity functions, rate matrices, or task representations rather than a common discrete vocabulary \citep{muller2022pfn,berghaus2026incontext,bae2023metatpp,verdenius2024latpfn}. Models for \glspl{des}, including CoLES, MOTOR, and PinFM, learn reusable representations within domains whose event schemas remain stable \citep{babaev2022coles,steinberg2024motor,chen2025pinfm}.

\paragraph{Event-sequence forecasting beyond a single stochastic-process family.} A separate line of work adapts a \emph{single} event-sequence model to changing data without our cross-alphabet, zero-shot goal. LAMP couples a TPP with a large language model performing few-shot abductive reasoning over retrieved historical events to improve mark prediction \citep{shi2023lamp}; HYPRO reweights an autoregressive TPP's proposals with a learned energy function for long-horizon, multi-step rollout \citep{xue2022hypro}; PromptTPP retrieves a continuous-time prompt pool to continually adapt a TPP to streaming, non-stationary event data \citep{xue2023prompttpp}. These target complementary problems: reasoning-augmented prediction, long-horizon calibration, and continual adaptation, respectively, within one dataset's fixed vocabulary, rather than a codebook shared, without any fine-tuning, across datasets whose event identities are mutually meaningless; we do not compare against them directly for this reason, but HYPRO in particular addresses the long-horizon rollout setting we flag as future work (Appendix~\ref{sec:app:appendix}).

\paragraph{Vocabulary symmetry and predictive discretization.} EquiTabPFN and lexinvariant language models address target-label and vocabulary permutations, respectively, but neither learns a shared predictive codebook across stochastic-process families \citep{arbel2025equitabpfn,huang2023lexinvariant}. Vector-quantized and predictive self-supervised methods learn discrete or latent representations whose alignment is inherited from a common visual or acoustic space \citep{oord2017vqvae,hsu2021hubert,assran2023ijepa,baek2025discretejepa}; predictive-state, causal-state, and information-bottleneck approaches instead motivate grouping histories by their future behavior \citep{singh2004psr,shalizi2001computational,creutzig2009pastfuture}. \gls{desjepa} connects these directions by learning a shared discrete prediction space across processes whose native event alphabets are unaligned.

\end{document}